\documentclass[a4paper,conference]{IEEEtran}
\usepackage{graphicx}
\usepackage{amsmath,amssymb} % define this before the line numbering.

\usepackage{float}

\usepackage[small]{caption}
\usepackage{wrapfig}
\usepackage{verbatim}

\ifCLASSINFOpdf
\else
\fi
\begin{document}
\title{Directed Variational Cross-encoder Network for Few-shot Multi-image Co-segmentation}

%\author{\IEEEauthorblockN{Sayan Banerjee}
%\IEEEauthorblockA{Department of Electrical Engineering \\ Indian Institute of Technology Bombay\\
%Email: sayan91.ban@gmail.com}
%\and
%\IEEEauthorblockN{S Divakar Bhat}
%\IEEEauthorblockA{Department of Electrical Engineering\\ Indian Institute of Technology Bombay\\
%Email: sdivakarbhat@gmail.com}
%\and
%\IEEEauthorblockN{Subhasis Chaudhuri\\ and Rajbabu Velmurugan}
%\IEEEauthorblockA{Department of Electrical Engineering\\ Indian Institute of Technology Bombay}}


\author{\IEEEauthorblockN{\IEEEauthorrefmark{1}Sayan Banerjee,
\IEEEauthorrefmark{1}S Divakar Bhat,
Subhasis Chaudhuri, and
Rajbabu Velmurugan
\IEEEauthorblockA{Department of Electrical Engineering\\ Indian Institute of Technology Bombay}
\IEEEauthorblockA{Email: sayan91.ban@gmail.com, sdivakarbhat@gmail.com, sc@ee.iitb.ac.in, rajbabu@ee.iitb.ac.in}
\IEEEauthorblockA{\IEEEauthorrefmark{1}\textit{denotes equal contribution}}}}
%Telephone: (800) 555--1212, Fax: (888) 555--1212}


\maketitle

\begin{abstract}
     In this paper, we propose a novel framework for multi-image co-segmentation using class agnostic meta-learning strategy by generalizing to new classes given only a small number of training samples for each new class. We have developed a novel encoder-decoder network termed as DVICE (Directed Variational Inference Cross Encoder), which learns a continuous embedding space to ensure better similarity learning. We employ a combination of the proposed DVICE network and a novel few-shot learning approach to tackle the small sample size problem encountered in co-segmentation with small datasets like iCoseg and MSRC. Furthermore, the proposed framework does not use any semantic class labels and is entirely class agnostic. Through exhaustive experimentation over multiple datasets using only a small volume of training data, we have demonstrated that our approach outperforms all existing state-of-the-art techniques. 
\end{abstract}


% \fi
\IEEEpeerreviewmaketitle

\section{Introduction}
\begin{figure}[t]
\centering
\includegraphics[width=1\linewidth]{ 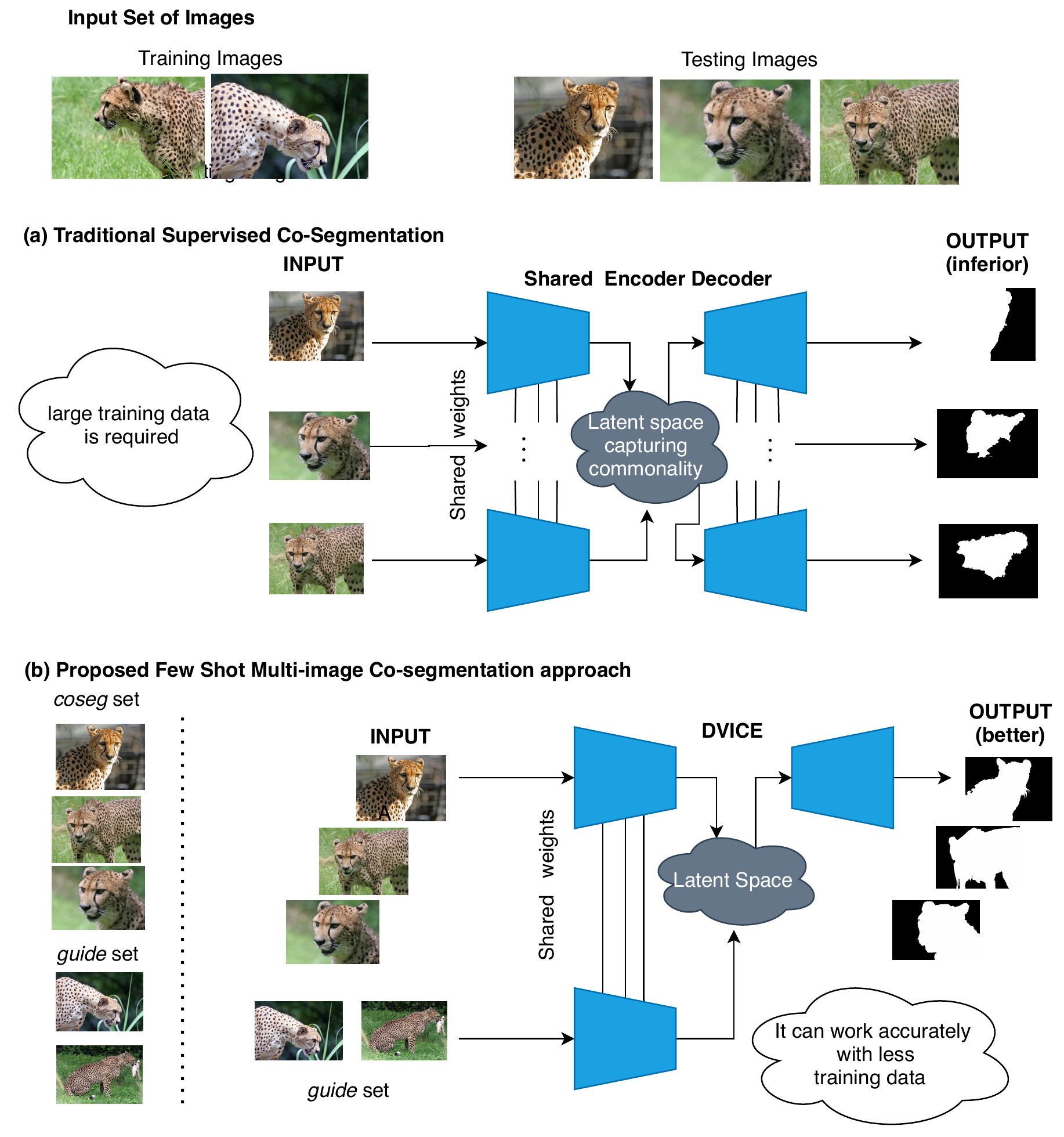}
    \captionof{figure}{A comparative illustration of the traditional supervised common foreground extraction approach where it expects a large set of training samples to learn commonality and to generate masks, hence fails for smaller training set as shown in (a). Whereas the proposed model using the same smaller training set with the help of a novel episodic training scheme performs better during testing. It is shown in (b)}
      \label{fig:motivation}
\end{figure}

Image co-segmentation is a fundamental problem in vision that aims to jointly identify and segment common objects from a set of images \cite{faktor2013co,rubinstein2013unsupervised}. In semantic image segmentation, each input is segmented independently into various meaningful regions along with the corresponding semantic labels. On the contrary, co-segmentation involves a group of similar input images to capture foreground segments of interest shared across them from as shown in Figure~\ref{fig:motivation}(a). To extract the shared foreground from each image, all images in the group are used to recognize commonality across the group. Such information aggregation across the group further helps to identify the common foreground when it is occluded or visually ambiguous from the background due to clutter. Hence, unlike semantic segmentation where semantics are used, co-segmentation algorithms are semantic-agnostic and learn commonality across images. Image co-segmentation has applications in image retrieval, annotation, object detection, and person re-identification.

Typical challenges in co-segmentation problem are: (i) to determine appropriate and consistent foreground features so that they can be detected when the object's appearance, shape, and pose vary significantly across input images, (ii) foregrounds having high similarity with background, and (iii) images that may not have a common foreground. 

\begin{figure*}
\begin{center}
\includegraphics[width=16cm,height=7.5cm]{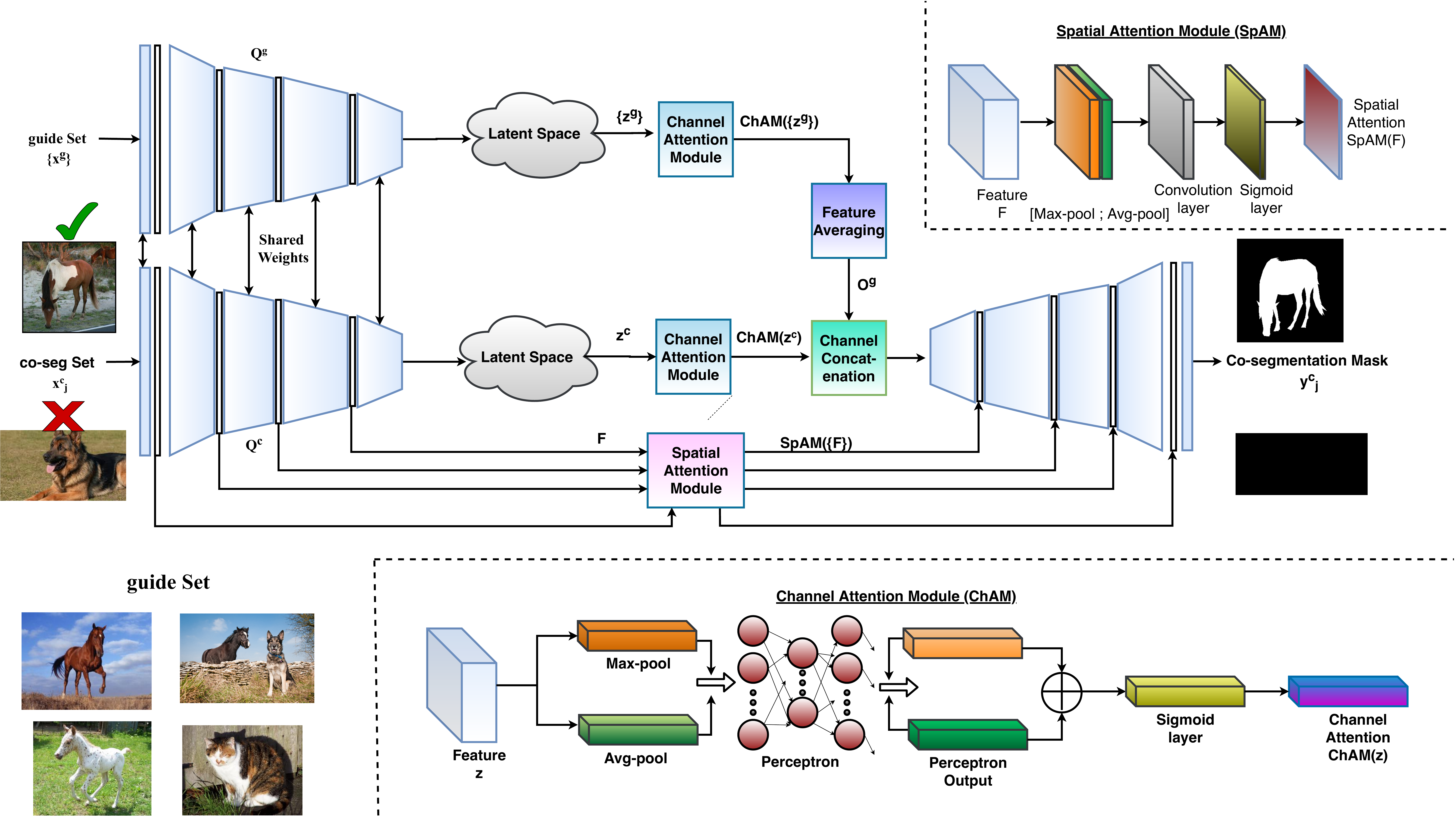}
\end{center}
  
   \caption{Illustration of the proposed pipeline for co-segmentation. \textbf{[Top-right]} The Spatial Attention Module (SpAM). \textbf{[Centre]} The complete pipeline showing the DVICE along with the SpAM and ChAM modules. \textbf{[Bottom]} The Channel Attention Module (ChAM). The detailed working of the individual modules is explained in Section~\ref{Arch}. Note that commonality corresponding to the foreground \textit{Horse} is learned from the \textit{guide} set,  which is exploited to segment the foreground of interest from the \textit{co-seg} set images. To investigate the robustness of our proposed method, we also experiment with \textit{guide} sets with outlier presence as illustrated.}

\label{fig:pipeline}
\end{figure*}

Recently, researchers have developed convolutional neural network (CNN) based models that automatically compute suitable features for co-segmentation with varying levels of supervision \cite{li2018deep,li2019group,yuan2017deep,chen2018semantic,banerjee2019cosegnet,li2019group}. However, all these methods require a large number of training samples for better feature computation and mask generation of the common foreground as illustrated in Figure~\ref{fig:motivation}(a). 
\begin{figure*}
\begin{center}
\includegraphics[height=7cm]{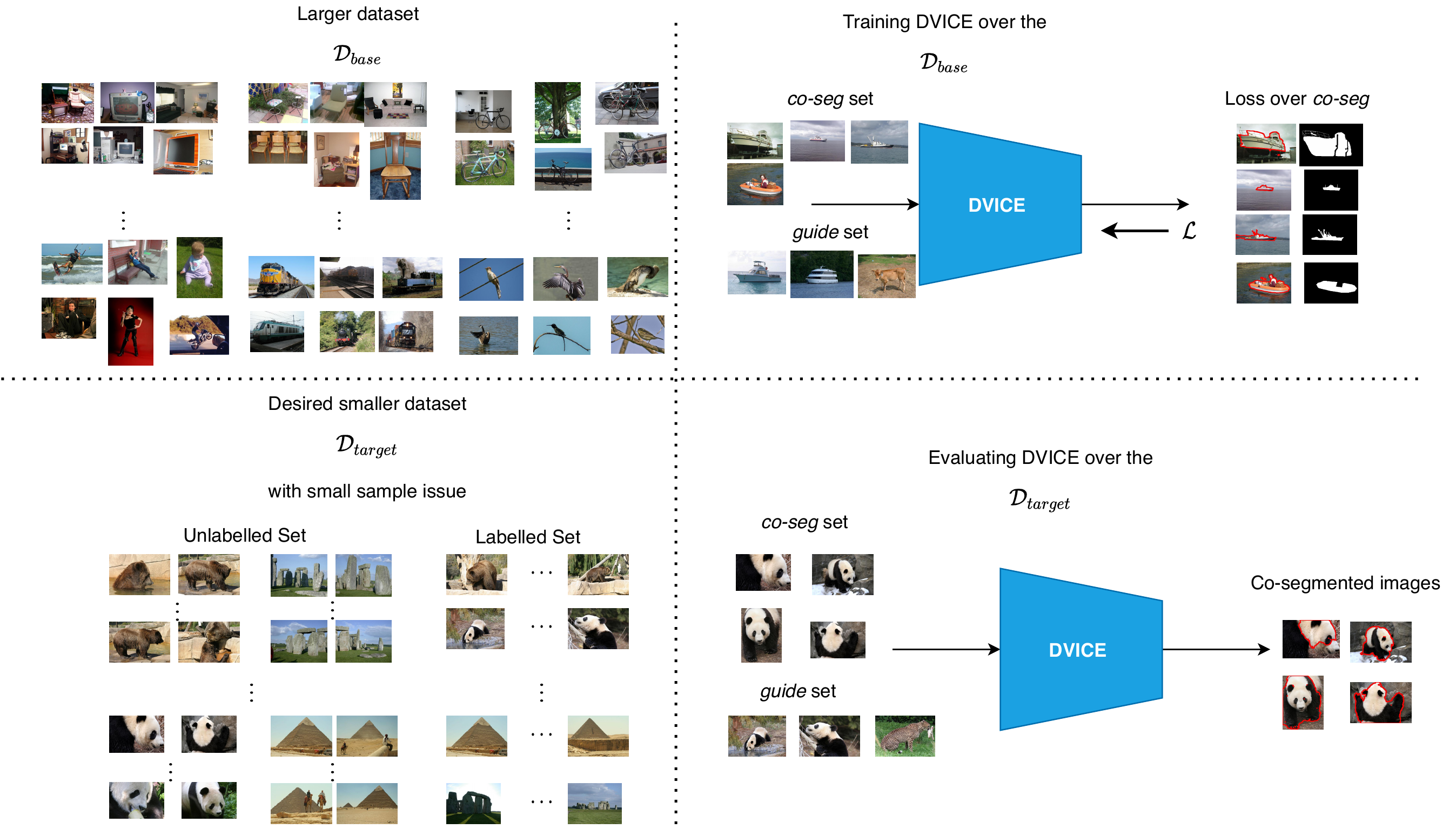}
\end{center}
  
   \caption{Illustration of the proposed approach using DVICE denoting the stages involved with the corresponding datasets $\mathcal{D}_{base}$ and $\mathcal{D}_{target}$, respectively. Note that the \textit{co-seg} set and \textit{guide} set shown here denote those for only an iteration out of many such combinations for the training and evaluation phases, respectively.}

\label{fig:method}
\end{figure*} 
Traditional supervised co-segmentation approaches require a large amount of annotated datasets while in many real scenarios we are presented with datasets with only a few labeled samples. Annotating input images in the form of a mask for the common foreground is also a very tedious task. Motivated by these challenges, in this paper, we investigate the problem of image co-segmentation in a few-shot setting. This implies performing a co-segmentation task over a set of a variable number of input images (called \textit{co-seg} set) by relying on the guidance provided by a set of images (called \textit{guide} set) to learn the commonality of features as shown in Figure~\ref{fig:motivation}(b). The proposed method learns commonality corresponding to the foreground of interest without any semantic information, for example as shown in Figure~\ref{fig:pipeline}, commonality corresponding to the foreground \textit{Horse} is learned from the \textit{guide} set,  which is exploited to segment the foreground of interest from the \textit{co-seg} set images. To investigate the robustness of our proposed method, we also experiment with a noisy \textit{guide} set (please see Figure~\ref{fig:pipeline}).

We propose a meta learning-based novel training method %solution 
where the proposed model learns the concept of co-segmentation %task
using a set of episodes sampled from a larger dataset, and subsequently adapts its knowledge to a smaller target dataset of new classes. Each episode consists of a \textit{guide} and \textit{co-seg} set that together mimic the few-shot scenario encountered in the smaller dataset. The proposed \textit{guide} set learns commonality using a novel, simple and robust feature integration technique and associates it with the \textit{co-seg} set individuals with the help of a variational encoder and an attention mechanism to segment the foreground of interest. The proposed encoder along with the attention mechanism helps to model the common foreground, where the intelligent feature integration method boosts the quality of its feature.  To improve the generalization capacity of the proposed end-to-end model, it is trained only using the co-segmentation loss computed over the \textit{co-seg} set.

Key contributions of this paper are: (i) We propose a novel multi-image co-segmentation framework capable of handling the \textbf{small sample size problem} and robust to outliers. (ii) We introduce a novel encoder-decoder network termed \textbf{DVICE: Directed Variational Inference Cross Encoder}, capable of performing few-shot learning explicitly for the co-segmentation task.
\section{Related Work}
\textbf{Co-segmentation:} Before the emergence of deep learning in co-segmentation, state-of-the-art works used graph matching \cite{ren2018mutual} and graph-based optimizations \cite{han2017robust, tsai2018image} with additional saliency cues \cite{han2017robust,ren2018mutual,tsai2018image,hsu2018co}. Subsequently, with deep learning, \cite{yuan2017deep} used object proposals but, their method is not end-to-end. Recently, \cite{chen2018semantic,li2018deep,banerjee2019cosegnet} proposed deep siamese encoder-decoder based end-to-end networks for the same. But, they can not handle more than two images or outliers efficiently. Li \textit{et al.} \cite{li2019group} developed a novel recurrent neural network (RNN) to perform co-segmentation over a variable number of images, but it fails when the input image set contains a significant number of outliers and their performance depends upon the input image sequence.\\
\textbf{Attention model:}
Attention modeling based techniques \cite{wang2017residual,hu2018squeeze} have become quite popular recently as they try to mimic human perception by focusing on salient features. \cite{woo2018cbam} uses both spatial and channel attention to suppress or emphasize intermediate features effectively. In this paper, we use attention modules to weigh the features and emphasize on the foreground of interest to extract consistent foreground from multiple images.\\
\textbf{Few-shot learning:} Recently few-shot learning techniques are applied widely in vision-based deep learning to tackle small sample size problem.  \cite{snell2017prototypical} showed promising results by introducing a prototypical network to learn an embedding space. \cite{kim2019variational} use prototypical learning along with variational inference technique to learn a continuous embedding space.\\
\textbf{Few-shot semantic segmentation:} These methods classify pixels of the \textit{query} image that belong to a target object class, conditioned by the ground-truth segmentation masks of a few \textit{support} images. Different from few-shot semantic segmentation methods \cite{dong2018few,shaban2017one,wang2019panet,zhang2019canet,zhang2019pyramid,siam2019amp,nguyen2019feature}, the proposed co-segmentation framework does not use any semantic masks explicitly to identify and compute features of the foreground of interest.

The proposed method exploits a few-shot learning approach to tackle the small sample size problem to perform multi-image co-segmentation. We introduce a conditional encoder-decoder framework that uses a mix of variational inference and meta-learning techniques specifically developed for the multiple-image co-segmentation task. To the best of our knowledge, this is the first work to use few-shot learning in a co-segmentation setting.

\section{Proposed Framework}
\subsection{Problem Setting}
Given a dataset $\mathcal{D}_{target} = \{(\widetilde{x}_i^t,\widetilde{y}_i^t)\}_{i=1}^n \cup \{\widetilde{x}_j^u\}_{j=1}^r$ containing a
small set of annotated training images and the corresponding ground truth masks $\{(\widetilde{x}_i^t,\widetilde{y}_i^t)\}_{i=1}^n$, our objective is to estimate co-segmentation masks $\{\widetilde{y}_1^u,\widetilde{y}_2^u,...,\widetilde{y}_r^u\}$ for the unlabeled target samples or the test set, $\{\widetilde{x}_1^u,\widetilde{x}_2^u,...,\widetilde{x}_r^u\}$. For this we follow a meta-learning approach, which is explained next.

\subsection{Few-shot Learning}
\label{fsl}
We define few-shot learning for co-segmentation as follows (inspired from~\cite{snell2017prototypical}): Formally we consider two datasets: a base set $\mathcal{D}_{base}$ with a large number of annotated samples and a target set $\mathcal{D}_{target}$ with a small number of annotated samples for co-segmentation respectively. We iteratively train our model over $\mathcal{D}_{base}$ using a series of episodes consisting of a \textit{guide} set and a \textit{co-seg} set. Each \textit{guide} and \textit{co-seg} set is designed such that it mimics the characteristics of the training and test set of $\mathcal{D}_{target}$ as shown in Figure~\ref{fig:motivation} and Figure~\ref{fig:method}. The role of the \textit{guide} and the \textit{co-seg} set is similar to to the support and query set typically encountered in the contemporary few-shot learning work \cite{snell2017prototypical, siam2019amp, nguyen2019feature}. %Precisely, At each iteration of training over the dataset $\mathcal{D}_{base}$ and on subsequent finetuning by using very few available labeled samples of $\mathcal{D}_{target}$,  we create \textit{co-seg} set $\mathcal{C}$ and \textit{guide} set $\mathcal{G}$, which is similar to the query set and support set split encountered in the contemporary few-shot learning work \cite{snell2017prototypical, siam2019amp, nguyen2019feature}. 
But unlike the support set used in those few-shot learning methods, the proposed \textit{guide} set does not rely on any semantic class labels and it is even tolerant to the presence of outliers while guiding the network to learn and perform extraction of the foreground of interest over the \textit{co-seg} set as shown in Figure~\ref{fig:pipeline}. It also serves as fine control over the foreground extraction process while testing over the $\mathcal{D}_{target}$ as can be seen in the figure~\ref{fig:co-seg for diff guide}. 
The \textit{guide} set discussed in this work includes samples (images) that contain a dominant class and samples (outlier images) that contain other non-dominant classes which we call positive and negative samples respectively. The positive samples share a common foreground the same as the foreground of interest that is to be extracted from the \textit{co-seg} set.

Due to the lack of sufficient training samples in the target dataset $\mathcal{D}_{target}$, we rely on meta-learning to learn and extract transferable embedding, thus facilitating us to learn and perform better foreground extraction on the target dataset. The proposed episodic training scheme is described in detail in the next section.

\subsection{Class Agnostic Episodic Training Scheme}
\label{episodic}
 We utilize a novel few-shot learning strategy to improve co-segmentation performance on the smaller target dataset $\mathcal{D}_{target}$, for which standard training leads to over-fitting. As mentioned before, We find a larger dataset denoted as $\mathcal{D}_{base}$ developed for the co-segmentation task to simulate the training scenario of $\mathcal{D}_{target}$, by creating multiple episodes over the $\mathcal{D}_{base}$. It is to be noted that we assume $\mathcal{D}_{target} \cap \mathcal{D}_{base} = \phi$. We develop an episodic training scheme, such that the proposed model learns to handle the co-segmentation task with few training samples without overfitting. 
 
Each episode consists of a \textit{guide} set and a \textit{co-seg} set such that the operation over $\mathcal{C}$ is directed by the set $\mathcal{G}$ as it provides the information of the common object to the \textit{co-seg} set $\mathcal{C}$  over which co-segmentation is performed. The \textit{guide} set $\mathcal{G}$ can even consist of noisy samples. Thus, the \textit{guide} set is designed as $\mathcal{G} = \{\mathcal{P}^{g} \cup \mathcal{N}^{g}\} = \{(x_1^g,y_1^g), \dots , (x_k^g,y_k^g) \}$, consisting of $n$~randomly selected positive samples $\{\mathcal{P}^{g}\}$ and $k-n$ randomly selected negative samples $\{\mathcal{N}^{g}\}$ and \textit{co-seg} set is $\mathcal{C}
= \{(x_1^c,y_1^c), \dots , (x_m^c,y_m^c) \}$. Here, $n$~is the number of annotated positive samples available in $\mathcal{D}_{target}$ and $k$~is the total number of samples in $\mathcal{G}$. \iffalse$n>(0.5\times k)$, the number of positive samples are only marginally higher than the negative samples.\fi Next with the help of the Encoder $Q$, which is a part of the proposed Directed Variational Inference Cross Encoder (DVICE), and channel attention (\text{ChAM}), the following feature averaging operation removes the influence of outliers and determines robust features $\mathcal{O}^g$ of the common object. The ChAM module is used to focus on the semantically meaningful part of the image by exploiting the inter-channel relationship of features. 

\begin{equation}
\begin{split}
 \mathcal{O}^g &= \frac{1}{|\mathcal{G}|} \sum_{i=1}^k \text{ChAM}(E(x_i^g))=\frac{1}{|\mathcal{G}|} \sum_{i=1}^k \text{ChAM}(z_i^g).    
\end{split}
\end{equation}
We refer to $\mathcal{O}^g$ as the common object prototype while $z_j^c$ and $z_j^g$ are the features obtained from the encoder for $j^{th}$ image of the \textit{co-seg} set and \textit{guide} set, respectively. The operation of $E$ and \text{ChAM} are explained in detail in section \ref{DVICE}, \ref{E-D} and \ref{ChAM}, respectively.\par
After computing $\mathcal{O}^g$, feature of individual samples $x_j^c\in\mathcal{C}, j=1 \dots m$ is obtained as,
\begin{equation}
 z_j^c=\text{ChAM}(E(x_j^c)).
\end{equation}
The $z_j^c$ so obtained are concatenated channel-wise with $\mathcal{O}^g$ and passed to the decoder.
The proposed decoder implicitly checks the similarity between the $\mathcal{O}^g$ and $z_j^c$, and estimates co-segmentation mask $\hat{y}_j^c$ for the input image $x_j^c$ accordingly. \par
The spatial importance of each pixel for specific layers of encoder features fed through the spatial attention module (SpAM) is also used by the decoder to localize the common foreground. Spatial attention aids the decoder to focus on the localization of the common foreground by providing a measure of importance for each pixel. 

While training for common foreground extraction, the proposed framework relies only on the assumption that there exists some degree of similarity between the \textit{guide} set and  \textit{co-seg} set. Thus we do not use any semantic class information during training as can be seen from Figure~\ref{fig:pipeline}, and hence, the proposed few-shot co-segmentation strategy is completely class agnostic.
\subsection{DVICE: Directed Variational Inference Cross Encoder}
\label{DVICE}
We propose a novel encoder-decoder model that is built on the theory of variational inference to learn a continuous feature space over input images for better generalization. But unlike the traditional variational auto-encoder setup, our proposed approach is a cross-encoder, mapping an input image $x^c$ to corresponding mask $y^c$ based on a directive $\mathcal{O}^g$ obtained from the \textit{guide} set.
Given the \textit{guide} set $\mathcal{G}$, an input image $x^c$ and the corresponding mask $y^c$ that are randomly sampled (where $(x^c,y^c)\in\mathcal{C}$) from an underlying unknown joint distribution $P(y^c,x^c;\theta)$, the purpose of the proposed encoder-decoder model is to estimate the parameters $\theta$ of the distribution from its likelihood. Thus, we are aiming to maximize the joint probability as shown below, 
\begin{equation}
    \max_{\theta}P(y^c,x^c) = \max_{\theta}\int_{z^c} \int_{\mathcal{O}^g}P(x^c,y^c,\mathcal{O}^g,z^c)d\mathcal{O}^gdz^c    
\label{joint}
\end{equation}
For clarity of presentation, we are dropping $\theta$ in $P(y^c,x^c)$. The process of finding the distribution $P(y^c,x^c)$ implicitly depends upon latent embedding of the sample $x^c$, which is $z^c$, and the common class prototype $\mathcal{O}^g$
computed over $\mathcal{G}$. The crux of the variational approach here is to learn the conditional distribution $P(z^c|x^c)$, that can produce the output mask $y^c$, and thus maximize $P(y^c,x^c)$. It is to be noted that $\mathcal{O}^g$ and $x^c$ that represents the common object prototype and a random sample from \textit{co-seg} set
$\mathcal{C}$ are independent of each other as the sets $\mathcal{G}$ and $\mathcal{C}$ are generated randomly. We proceed using an approach similar to the one in~\cite{blei2017variational} for deriving the Evidence Lower Bound (ELBO). From equation \eqref{joint},
\begin{equation}
\begin{split}
   P(y^c,x^c) & =\int_{z^c} \int_{\mathcal{O}^g} P(y^c|\mathcal{O}^g,z^c)P(\mathcal{O}^g|z^c,x^c)P(z^c|x^c)\\
            & \hspace{4.5cm}P(x^c)d\mathcal{O}^gdz^c \\
            & = \int_{z^c} \int_{\mathcal{O}^g} P(y^c|\mathcal{O}^g,z^c)P(\mathcal{O}^g)P(z^c|x^c)\\
            & \hspace{3cm}P(x^c) d\mathcal{O}^g dz^c \\
\end{split}
\label{eqn:variational}
\end{equation}
Since $z^c$ is the latent feature corresponding to $x^c$, we refrain from using them together inside joint probability as they provide redundant information. The main idea behind variational method used here is to learn distribution $Q(\mathcal{O}^g)$ and $Q(z^c|x^c)$ that can approximate the distributions $P(\mathcal{O}^g)$ and $P(z^c|x^c)$ over the latent variables, respectively.
Therefore, equation \eqref{eqn:variational} can be written as,
\begin{equation}
\begin{split}
 P(y^c,x^c) & = \int_{z^c} \int_{\mathcal{O}^g} \frac{P(y^c,\mathcal{O}^g,z^c)}{Q(\mathcal{O}^g,z^c)}Q(\mathcal{O}^g)Q(z^c|x^c)\\
                & \hspace{3cm}P(x^c)d\mathcal{O}^gdz^c\\
                & = P(x^c)\left[\mathbb{E}_{(\mathcal{O}^g,z^c) \sim Q(\mathcal{O}^g,z^c)} \frac{P(y^c,\mathcal{O}^g,z^c)}{Q(\mathcal{O}^g,z^c)}\right]\\
\end{split}
\label{eqn:expectation}
\end{equation}
Taking the log of equation \eqref{eqn:expectation} followed by Jensen's inequality,
\begin{equation}
\begin{split}
\log P(y^c,x^c) & \geq \mathbb{E}_{(\mathcal{O}^g,z^c) \sim Q(\mathcal{O}^g,z^c)} \log \frac{P(y^c,\mathcal{O}^g,z^c)}{Q(\mathcal{O}^g,z^c)}\\
               & \geq \mathbb{E}_{(\mathcal{O}^g,z^c) \sim Q(\mathcal{O}^g,z^c)}\left[\log P(y^c|\mathcal{O}^g,z^c)\right] \\
               & - KL\left[Q(\mathcal{O}^g|\mathcal{G}) || P(\mathcal{O}^g|\mathcal{G})\right]\\
               & - KL\left[Q(z^c|x^c) || P(z^c|x^c)\right]
\end{split}
\label{ELBO}
\end{equation}
From the ELBO obtained in equation \eqref{ELBO}, maximizing it will in turn result in the maximization of the target log-likelihood of generating a mask $y^c$ for a given input image $x^c$. Thus, unlike the traditional variational auto-encoders, here we learn to approximate a continuous embedding $Q$  which is capable of generating a mask $y^c$ given the input image $x^c$. The terms $Q(z^c|x^c)$ and $Q(\mathcal{O}^g)$ denote the mapping operation of encoder with shared weights and $P(y^c|\mathcal{O}^g,z^c)$ denotes the decoder part responsible for generating the co-segmentation mask given the common object prototype $\mathcal{O}^g$ and the latent embedding $z^c$.

We derive an empirical loss ($\mathcal{L}$) from equation \eqref{ELBO}, calculated over the \textit{co-seg} set, to train our model which is shown below,
\begin{equation}
\begin{split}
\mathcal{L} &= -\sum_{j=1}^m\sum_{(a,b)}\log P(y_j^c(a,b)|\mathcal{O}^g,z_j^c)\\
               & + KL\left[Q(\mathcal{O}^g|\mathcal{G}) || P(\mathcal{O}^g|\mathcal{G})\right]\\
               & + KL\left[Q(z^c|x^c) || P(z^c|x^c)\right]
\end{split}
\end{equation}
where $y_j^c(a,b)$ is the predicted label of the mask at the pixel location $(a,b)$. The model is trained over the larger dataset $\mathcal{D}_{base}$ using multiple episodes until convergence. 
%A detailed mathematical formulation and implementation strategy of the proposed DEVICE for example the reparameterization trick \cite{kim2019variational} are explained in the supplementary material.

To perform co-segmentation over $\mathcal{D}_{target}$,  $\{(\widetilde{x}_i^t\}_{i=1}^n$, is used as the \textit{guide} set and $\{\widetilde{x}_1^u,\widetilde{x}_2^u,...,\widetilde{x}_r^u\}$ is used as the \textit{co-seg} set. Hence, the final co-segmentation accuracy of proposed method is examined over the corresponding \textit{co-seg} set of $\mathcal{D}_{target}$.

\section{Network Architecture}
\label{Arch}
The proposed network architecture is shown in Figure~\ref{fig:pipeline}. ResNet-50 forms the backbone of the encoder-decoder framework used in this approach. The encoder-decoder framework in combination with the channel and spatial attention modules form the complete pipeline. Unlike \cite{hsu2018co}, which uses attention in cascade with the encoder, we implement attention for channel and spatial localisation of foreground as introduced in \cite{woo2018cbam} with ChAM over the channels of the feature and SpAM focusing on spatial localisation complementing the ChAM module. The individual modules building this framework as shown in Figure~\ref{fig:pipeline} are explained briefly in this section.
\subsection{Encoder-Decoder}
\label{E-D}
The variational encoder-decoder is a novel modification of the variational autoencoder network. The encoder-decoder structure is implemented using the ResNet-50 architecture at its backbone. The encoder ($E$) is just the ResNet-50 network with a final additional $1 \times 1$ convolutional layer. The decoder has five stages of up sampling and convolutional layers with skip connections through a spatial attention module as shown in Figure~\ref{fig:pipeline}. The encoder and decoder are connected through a channel attention module.
\subsection{Channel Attention Module (ChAM)}
\label{ChAM}
Both average-pooling and max-pooling are performed simultaneously on a feature map $z$ to boost the representational power of the network. The output from these operations $z_{\text{avg}}$ and $z_{\text{max}}$, respectively, are then fed to a perceptron $\Phi$ to produce the channel attention weights $W_c \in \mathbb{R}^{N_c\times1\times1}$, where $N_c$ is the number of channels. The output so obtained from the multi-layer perceptron is then added element-wise and passed through sigmoid as shown.
\begin{equation}
    W_c(z) = \sigma\left(\Phi(z_{\text{avg}})+\Phi(z_{\text{max}})\right).
\end{equation}
\subsection{Spatial Attention Module (SpAM)}
\label{SpAM}
The inter-spatial relationship among features is utilized to generate the spatial attention map. To generate the attention map for a given feature $F$, both average-pooling and max-pooling are applied across the channels, resulting in $F_{\text{avg}}$ and $F_{\text{max}}$, respectively these are concatenated to form $[F_{\text{avg}}; F_{\text{max}}]$. Convolution operation $f(.)$ followed by a sigmoid function is performed over the concatenated features to get a spatial attention map $W_s \in \mathbb{R}^{H \times W}$, where $H$ and $W$ represent the height and width of the feature map.
\begin{equation}
    W_s(F) = \sigma\left(f([F_{\text{avg}} ;F_{\text{max}}])\right).
\end{equation}

\section{Experimental Results}
For the proposed framework, we consider the Pascal-VOC dataset as the $\mathcal{D}_{base}$ over which we perform the class-agnostic episodic training as discussed in Section~\ref{episodic}. Following this, we consider three various datasets as $\mathcal{D}_{target}$: iCoseg, MSRC, and Internet datasets over which the model is then fine-tuned. The iCoseg and MSRC datasets are challenging due to the limited number of samples available in each of them, and not ideal for supervised learning.
Our proposed approach overcomes this small sample problem by using a few-shot learning method for training.

We evaluate the proposed method on the test set of co-segmentation datasets: iCoseg and MSRC and compare its performance with state-of-the-art methods using Precision ($\mathcal{P}$) and Jaccard Index ($\mathcal{J}$). Apart from the above datasets, we also experiment over the Internet dataset with a variable number of co-segmentable images along with outliers. Visual results on these datasets are presented for different sets of input images.
%, one set, in each row. Within each row, the image on left is the input image and that on the immediate right is the corresponding co-segmented image.
%-------------------------------------------------------------------------

\subsection{Implementation Details}
We use pre-trained ResNet-50 as our encoder. For, the rest of the network we follow \cite{glorot2010understanding} for initializing weights. For the optimization, we use stochastic gradient descent with the learning rate and momentum $1\times10^{-5}$ and $0.9$, respectively for all of the datasets. We resize each input image and the corresponding mask to $224 \times 224$ pixels and apply random rotation and horizontal flipping on them for augmenting the number of training samples. For all of the datasets, set $\mathcal{G}$ and set $\mathcal{C}$ are randomly created such that there are no common images and we use the episodic training scheme described in \ref{episodic}.

\subsection{Performance Comparison on Datasets}
\textbf{Pascal-VOC} \cite{faktor2013co} dataset consists of 20 different classes with 50 samples per class where samples within a class have significant appearance and pose variations. We consider this as our $\mathcal{D}_{base}$.

\textbf{iCoseg}~\cite{batra2010icoseg} dataset is a relatively smaller dataset which has 38 classes with 643 images. Some classes have less than 5 samples. Since, the number of labeled samples are small, we consider this dataset as one of our $\mathcal{D}_{target}$ dataset. It should be noted that the dataset is very small furthermore to examine our proposed few shot method, we split the dataset into training and testing set in the ratio of 1:1 and as a result the \textit{guide} set to \textit{co-seg} set ratio is also 1:1. We compare performance of our method with state-of-the-art methods. As seen in Table~\ref{table:icoseg} our method outperforms others at least by a margin of 5\% in $\mathcal{J}$. 
\begin{table}[!tb]
    \centering
   \begin{tabular}{|c|c|c|}
\hline
Method & Precision ($\mathcal{P}$) & Jaccard Index ($\mathcal{J}$) \\
\hline\hline
\cite{li2018deep} & - & 0.84\\\hline
\cite{ren2018mutual} & - & 0.73\\\hline
\cite{han2017robust} & 94.4 & 0.78\\\hline
\cite{chen2018semantic} & - & 0.87\\\hline
\cite{hsu2018co} & 96.5 & 0.77\\\hline
\cite{tsai2018image} & 90.8 & 0.72\\\hline
\cite{li2019group} & 97.9 & 0.89\\\hline
[ours] & \textbf{99.1} & \textbf{0.94}\\\hline
\end{tabular}
\caption{Comparison using iCoseg dataset.} \label{table:icoseg}
\end{table}

\begin{figure}[!bt]
    \centering
    \includegraphics[width=0.7\linewidth]{ 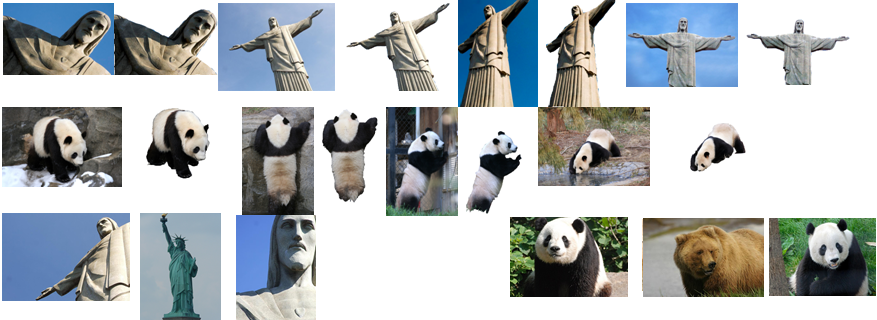}
    \captionof{figure}{Visual results of the proposed method evaluated on the iCoseg dataset. The first two rows depict the set of images used for co-segmentation (\textit{co-seg} set) with their corresponding results to the immediate right of each image. The last row denotes the \textit{guide} sets used to guide the network towards the desired foreground. The first three images correspond to the \textit{guide} set for the first row, while the last three images from the last row correspond to the \textit{guide} set of the second row of images. Note that the model is robust to the presence of outliers/noise in the  \textit{guide} sets as can be seen in the \textit{guide} set corresponding to the \textit{Panda}.}
    \label{fig:icoseg}
\end{figure}

It should be noted that none of the other methods can exploit the small number of labeled samples of the iCoseg dataset, whereas with the proposed few-shot learning scheme we can fine-tune our model over the small set of available samples without any overfitting, which inherently boosts our performance. The method in \cite{li2019group} created additional annotated data to tackle the small sample size problem, which essentially requires extra human supervision. Visual results are shown in Figure~\ref{fig:icoseg}. It can be seen that our method performs well even for the most difficult class (\textit{Panda}).

\begin{figure}[!tb]
    \centering
   \includegraphics[width=1\linewidth]{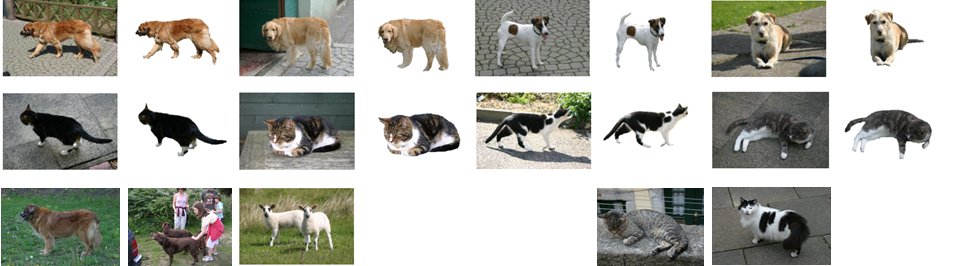}
    \captionof{figure}{Visual results on the MSRC dataset with first two rows depicting the set of images used for co-segmentation (\textit{co-seg} set) with their corresponding results to the immediate right of each image. The last row denotes the \textit{guide} sets used to guide the network towards the desired foreground. The first three images correspond to the \textit{guide} set for the first row, while the last two images from the last row correspond to the \textit{guide} set of the second row of images.}
    \label{fig:msrc}
\end{figure}

\textbf{MSRC}~\cite{vicente2011object} dataset consists of the following classes: \textit{cow}, \textit{plane}, \textit{car}, \textit{sheep}, \textit{cat}, \textit{dog} and \textit{bird}. Each class has 10 images. We remove the aforementioned 7 classes from  $\mathcal{D}_{base}$ (Pascal-VOC) to preserve the few-shot setting in our experiment. We use the training and testing split as 2:3. The quantitative and visual results are shown in Table~\ref{tab:msrc} and in Figure~\ref{fig:msrc}.  
\begin{table}[!bt]
\centering
 \begin{tabular}{|c|c|c|} \hline
 Method & Precision $\mathcal{P}$ & Jaccard Index $\mathcal{J}$ \\ \hline\hline
  \cite{faktor2013co} & 92.0 & 0.77 \\\hline
  \cite{li2018deep} & 94.4 & 0.80 \\\hline
 \cite{chen2018semantic} & 95.3 & 0.77 \\\hline
 \cite{banerjee2019cosegnet} & 96.3 & 0.85\\ \hline
 [ours] & \textbf{98.7} & \textbf{0.88}\\\hline 
\end{tabular}
\captionof{table}{Comparison of methods on the MSRC dataset.}
\label{tab:msrc}
\end{table}
However, it can be seen from Table~\ref{tab:msrc} that the most competitive method \cite{banerjee2019cosegnet} performs co-segmentation only over two images and use a train to test split as 3:2 but the proposed method still outperforms it by a margin of 3\%.

\begin{figure}[tb!]
    \centering
\includegraphics[width=1\linewidth]{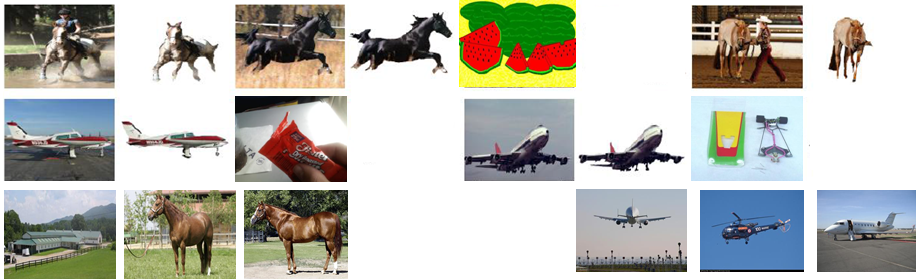}
    \captionof{figure}{Visual results of the proposed method evaluated on the Internet dataset. The first two rows depict the set of images used for co-segmentation (\textit{co-seg} set) with their corresponding results to the immediate right of each image. The last row denotes the \textit{guide} sets used to guide the network towards the desired foreground. The first three images correspond to the \textit{guide} set for the first row, while the last three images from the last row correspond to the \textit{guide} set of the second row of images. Note that the model is robust to the presence of outliers in the  \textit{guide} sets as can be seen in the \textit{guide} set corresponding to the horse. Also, blank white results denote no foreground extraction for the images with the desired foreground absent.}
      \label{fig:Internet}
\end{figure}

\textbf{Internet}~\cite{rubinstein2013unsupervised} dataset has 3 classes namely \textit{car}, \textit{Aeroplane} and \textit{Horse} with 100 samples per class. Though the number of classes is small, this dataset has high intra-class variation and is relatively large. But to examine the performance of our proposed few-shot method, we split it as per the ratio of 1:9 into training and testing set whereas the method in \cite{banerjee2019cosegnet} used a ratio of 3:2. As we consider Pascal-VOC as $\mathcal{D}_{base}$, we remove the above three classes from it. For the experimentation, we vary the number of images of the \textit{co-seg} set (randomly selected 40, 60, or 80 images from the Internet dataset) and also vary the number of outliers from 10\% to 50\% of the total sample of the set in steps of 10. We report the average accuracy computed over all of these sets in Table~\ref{table:Internet}. 
 \begin{table}[tb!]
     \centering
\begin{tabular}{|c|c|c|}
\hline
Method & Precision ($\mathcal{P}$) & Jaccard Index ($\mathcal{J}$) \\
\hline\hline
 \cite{li2018deep} & 93.3 & 0.70\\\hline
\cite{ren2018mutual} & 85.0 & 0.53\\\hline
\cite{chen2018semantic} & - & 0.74\\\hline
\cite{hsu2018co} & 92.2 & 0.69\\\hline
\cite{banerjee2019cosegnet} & 96.1 & 0.77\\\hline
\cite{li2019group} & 97.1 & 0.84\\\hline
[ours] & \textbf{99.0} & \textbf{0.87}\\\hline
\end{tabular}
\caption{Comparison using Internet dataset.}
\label{table:Internet}
\end{table}
This shows that our method can handle large number of input images and also large number of outliers. The visual results are shown in Figure~\ref{fig:Internet}. 

\subsection{Ablation Study}

\begin{figure}[!tb]
\begin{center}
%\fbox{\rule{0pt}{2in}\includegraphics[width=1\linewidth]{Figures/icoseg_results.png} \rule{.9\linewidth}}
\includegraphics[width=0.9\linewidth]{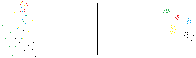} 
\end{center}
   \caption{Comparison of the embedding space obtained by the proposed DVICE setup (right) and the ResNet-50 based encoder (left) using t-SNE plots. Each class is represented by one color.}
\label{fig:tsne}
\end{figure}

The task of image co-segmentation can be divided into two sub-tasks in cascade. The first task is to identify similar objects without exploiting any semantic information or more formally cluster similar objects together. The second task is to jointly segment similar objects or performing foreground segmentation over each cluster. In this context, to show the role of the proposed Directed Variational Inference Cross Encoder (DVICE) for clustering, we replace the proposed encoder with the ResNet50 of which the final two layers are removed. We compare the embedding space obtained with the normal ResNet50 based encoder and the proposed DVICE using t-SNE plots in Figure~\ref{fig:tsne}. We run the experiment on the MSRC dataset where we randomly choose 5 classes to examine the corresponding class embedding. As can be seen the proposed encoder with the help of variational inference reduces intra-class distances and increases inter-class distances implicitly, which in turn boosts the co-segmentation performance, significantly.

\begin{figure}[!tb]
\begin{center}
%\fbox{\rule{0pt}{2in}\includegraphics[width=1\linewidth]{Figures/icoseg_results.png} \rule{.9\linewidth}}
\includegraphics[width=1\linewidth]{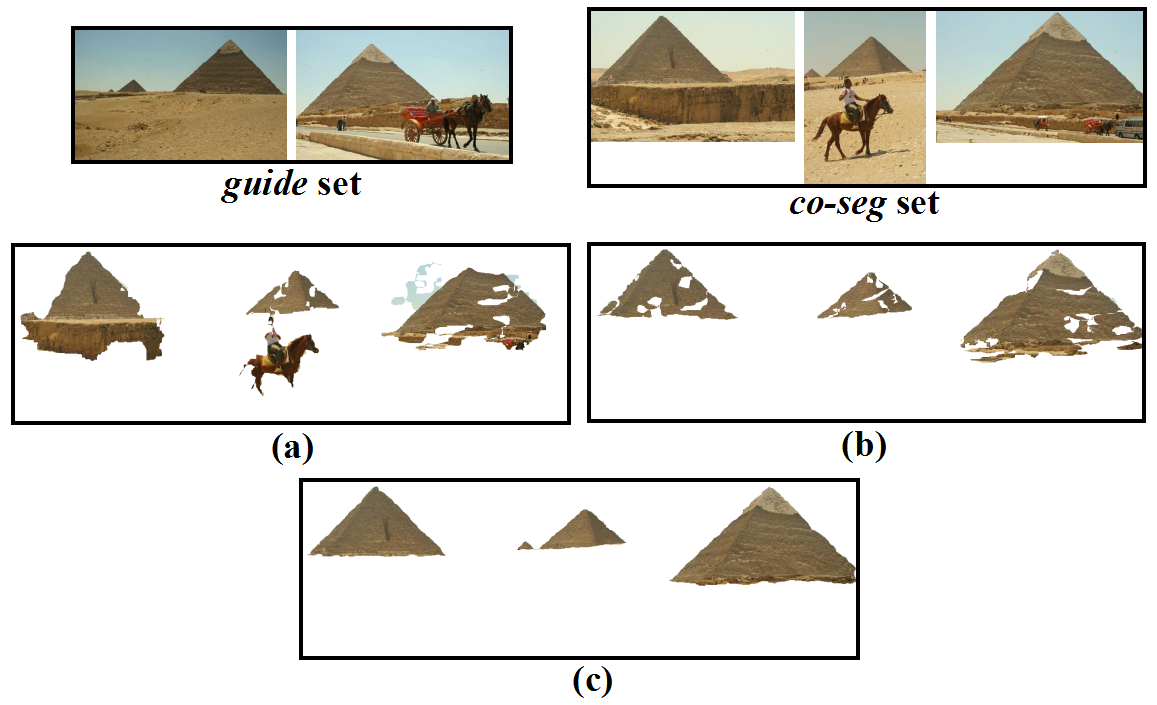} 
\end{center}
   \caption{Visual results for co-segmentation  \textbf{(a)} Output without ChAM and SpAM . \textbf{(b)} Output with ChAM but without SpAM. \textbf{(c)} Output with ChAM and SpAM modules. Note that attention properly highlights the pyramid, the common foreground in the \textit{guide} set, resulting in extraction of correct foreground and ignores other foregrounds present in the \textit{co-seg} set. Whereas the model is confused in the absence of attention and results in extraction of incorrect foreground.}
\label{fig:attention}
\end{figure}

The proposed channel attention module (ChAM) and spatial attention module (SpAM) also play a significant role to obtain the common object in the input image set. As can be seen from Figure~\ref{fig:attention}(a) and Figure~\ref{fig:attention}(c), ChAM and SpAM help the proposed model to identify common objects in a very cluttered background and objects with different scales. However, the role of the ChAM is more crucial to identify common objects whereas the SpAM is responsible for better mask production. Therefore, as can be seen from the Figure~\ref{fig:attention}(b), although the proposed model can identify the common object without the SpAM, it generates spurious output.

\begin{figure}[!tb]
\begin{center}
%\fbox{\rule{0pt}{2in}\includegraphics[width=1\linewidth]{Figures/icoseg_results.png} \rule{.9\linewidth}}
\includegraphics[width=0.8\linewidth]{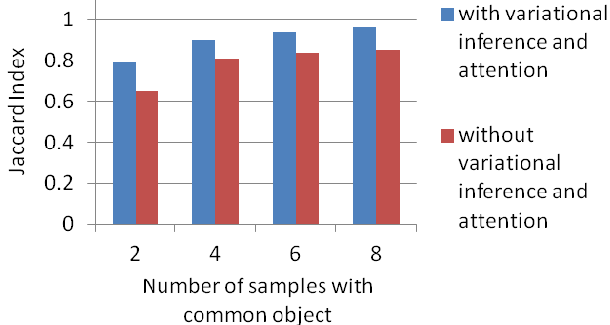} 
\end{center}
  
   \caption{Jaccard Index ($\mathcal{J}$) with varying number of positive samples in the \textit{guide} set. This denotes the comparison between the Jaccard Index with and without variational inference and attention with the former dominating in performance. Also, note that the model shows appreciable performance even with a smaller number of positive samples in the \textit{guide} set.}
\label{fig:guide}
\end{figure}

We calculate the common object prototype, $\mathcal{O}^g$, from the set $\mathcal{G}$ by feature averaging. It can be seen that the method of determining $\mathcal{O}^g$ is similar to noise cancellation where the motivation is to reduce the impact of outliers and to increase the influence of the positive samples (samples containing the common object). We experiment on the iCoseg dataset where we vary the number of positive samples in the \textit{guide} set to be 2, 4, 6, 8. The size of the \textit{guide} set is fixed at 8. The performance of the proposed method with and without the proposed variational inference and the attention modules is shown in Figure~\ref{fig:guide}. It can be seen that the proposed method is robust against outliers and can work with a small number of positive guide samples.

\begin{figure}[!tb]
\begin{center}
%\fbox{\rule{0pt}{2in}\includegraphics[width=1\linewidth]{Figures/icoseg_results.png} \rule{.9\linewidth}}
\includegraphics[width=1\linewidth]{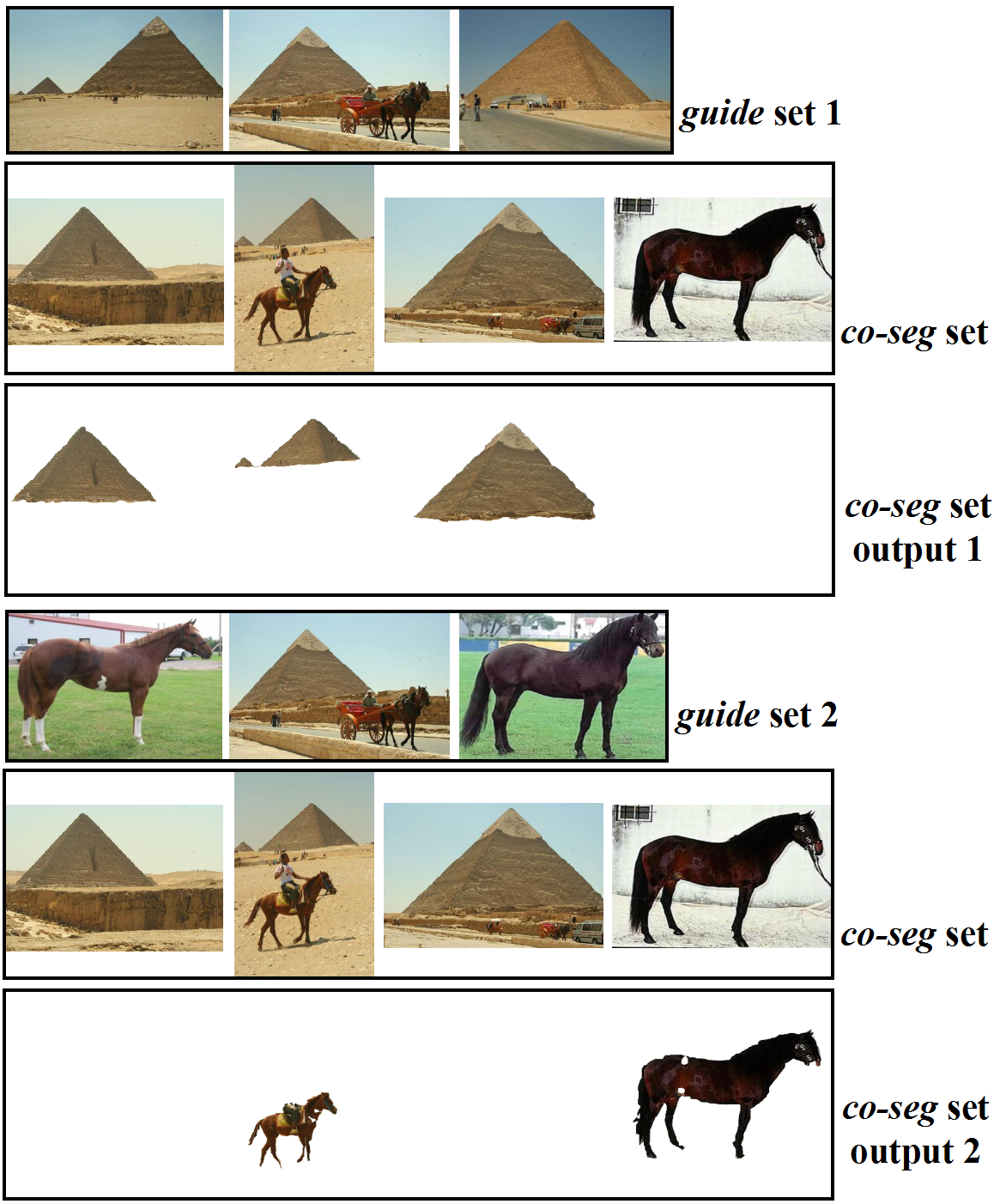} 
\end{center}
   \caption{Visual results of the proposed method over one \textit{co-seg} set but with different \textit{guide} sets. Note that we are able to achieve fine control over the foreground extraction by varying the composition of majority samples as seen in the \textit{guide} set 1 (where \textit{pyramid} is the majority) and \textit{guide} set 2 (where \textit{horse} is the majority), the corresponding outputs obtained over the \textit{coseg} set.}
\label{fig:co-seg for diff guide}
\end{figure}

We also demonstrate the fine control of our approach over the foreground extraction process as shown in Figure~\ref{fig:co-seg for diff guide}. Here, for a given \textit{co-seg} set with multiple, potential common foregrounds i.e., \textit{pyramid} and \textit{horse}, we are able to guide the network to perform foreground extraction on the \textit{co-seg} set for each of these foregrounds just by varying the composition of the \textit{guide} set.

\section{Conclusion}
We propose a new framework to perform multiple image co-segmentation, which is capable of overcoming the small-sample problem by integrating few-shot learning and variational inference. We have shown that our framework is capable of learning a continuous embedding to extract consistent foreground from multiple images of a given set. The introduced approach is capable of performing consistently, even in the presence of a large number of outlier samples in the \textit{co-seg} set. 
We have demonstrated that the proposed approach achieves state-of-the-art performance in co-segmentation over small datasets and have validated the same over multiple datasets.


%\section*{Acknowledgment}




\bibliographystyle{IEEEtran}
\bibliography{refs}
%\begin{thebibliography}{1}


%\end{thebibliography}

\end{document}